\documentclass[conference]{IEEEtran}
\IEEEoverridecommandlockouts

\usepackage{cite}
\usepackage{amsmath,amssymb,amsfonts}
\usepackage{algorithmic}
\usepackage{graphicx}
\usepackage{textcomp}
\usepackage{xcolor}
\def\BibTeX{{\rm B\kern-.05em{\sc i\kern-.025em b}\kern-.08em
    T\kern-.1667em\lower.7ex\hbox{E}\kern-.125emX}}

\usepackage{multirow}
\usepackage{algorithm}
\usepackage{array}
\usepackage[caption=false,font=normalsize,labelfont=sf,textfont=sf]{subfig}
\usepackage{stfloats}
\usepackage{url}
\usepackage{verbatim}

\newcommand{\x}{\mathbf{x}}
\newcommand{\z}{\mathbf{z}}

\newcommand{\zero}{\mathbf{0}}
\newcommand{\Id}{\mathbf{I}}

\newcommand{\Gauss}[2]{\mathcal{N}\left(#1,#2\right)}

\allowdisplaybreaks
\begin{document}


\title{Disentanglement Analysis in Deep Latent Variable Models Matching Aggregate Posterior Distributions}

\author{\IEEEauthorblockN{Surojit Saha}
\IEEEauthorblockA{\textit{SCI, Kahlert School of Computing} \\
\textit{The University of Utah}\\
Salt Lake City, USA \\
surojit.saha@utah.edu}
\and
\IEEEauthorblockN{Sarang Joshi}
\IEEEauthorblockA{\textit{SCI, Kahlert School of Computing} \\
\textit{The University of Utah}\\
Salt Lake City, USA \\
sarang.joshi@utah.edu}
\and
\IEEEauthorblockN{Ross Whitaker}
\IEEEauthorblockA{\textit{SCI, Kahlert School of Computing} \\
\textit{The University of Utah}\\
Salt Lake City, USA \\
whitaker@cs.utah.edu}
}

\maketitle

\begin{abstract}
Deep latent variable models (DLVMs) are designed to learn meaningful representations in an unsupervised manner, such that the hidden explanatory factors are interpretable by independent latent variables (aka disentanglement). The variational autoencoder (VAE) \cite{kingma2014auto,rezende2014stochastic} is a popular DLVM widely studied in disentanglement analysis due to the modeling of the posterior distribution using a factorized Gaussian distribution\cite{VAERecPCA_CVPR_2019} that encourages the alignment of the latent factors with the latent axes. Several metrics have been proposed recently, assuming that the latent variables explaining the variation in data are aligned with the latent axes (cardinal directions). However, there are other DLVMs, such as the AAE and WAE-MMD (matching the aggregate posterior to the prior), where the latent variables might not be aligned with the latent axes. In this work, we propose a statistical method to \emph{evaluate disentanglement} for any DLVMs in general. The proposed technique discovers the latent vectors representing the generative factors of a dataset that \emph{can be different from the cardinal latent axes}. We empirically demonstrate the advantage of the method on two datasets.
\end{abstract}


\section{Introduction}
\emph{Deep latent variable models} (DLVMs) have gained a great deal of well-deserved attention due to their ability to \emph{model the distribution} of the high-dimensional, complex datasets \cite{score_based_diffusion,avae_saha_2023,ard_vae_saha_2025} and learn meaningful representations \cite{Repr_learn_PAMI_2013,tschannen2018recent,gpunet_2022,multitask_saha_2023,Xiwen_real_time_idling_2024,Xiwen_joint_audio_visual_idling_2024} for downstream applications, such as \emph{few-shot learning} \cite{Model_FSL_Laro_2017,MAML_FSL_Finn_2017,gpunet_2022}. 
DLVMs learn a joint distribution distribution, $p_{\theta}(\x{}, \z{})$, that captures the relationship between a set of learned, hidden variables, $\z{}$, and the observed variables, $\x{}$. The variational autoencoder (VAE) \cite{kingma2014auto,rezende2014stochastic} is a popular DLVM. Learning \emph{disentangled representations} in an unsupervised framework is a desired property of a DLVM such that independent latent variables can explain the variability in the observed data \cite{Repr_learn_PAMI_2013}. DLVMs learn disentangled representations by encoding meaningful information onto the independent latent variables, such that each latent variable, $\z{}_i$, represents \emph{only} a single generative factor of the data; thus, making $\z{}_i$ \emph{interpretable}.

The DLVM uses an \emph{encoder-decoder} architecture, where the encoder projects the observed data onto a low dimensional manifold, and the decoder reconstructs the encoded representations.
The encoded representations are mapped to a prior distribution, and typically, standard normal distribution, $\Gauss{\zero}{\Id}$, is chosen as the prior distribution in DLVMs. $\Gauss{\zero}{\Id}$ is invariant to rotations, implying there is no difference in the expressiveness of a latent sample, $\z{}$, and its rotated counterpart, $rot(\z{})$, in terms of the reconstruction by the decoder. However, the latent representations are not interpretable anymore on rotation \cite{VAERecPCA_CVPR_2019,Dis_benchmark_ICML_2019}. Despite this limitation, the VAE (using $\Gauss{\zero}{\Id}$ as the prior) succeeds in learning a disentangled representation due to the modeling of the posterior distribution in the latent space with a factorized Gaussian distribution\cite{VAERecPCA_CVPR_2019}. This motivated the development of the different variants of the VAE that encourage the learning of disentangled representations \cite{beta_VAE_ICLR_2017,Factor_VAE_ICML_2018,TC_VAE_Neurips_2019,ard_vae_saha_2025}. In addition, a slew of metrics were proposed to evaluate the disentanglement of the learned representations \cite{beta_VAE_ICLR_2017,Factor_VAE_ICML_2018,TC_VAE_Neurips_2019,DIP_VAE_ICLR_2018,DCI_ICLR_2018,DCI_ES_ICLR_2023}.

Increasing the strength of the regularization loss in VAEs for improved disentanglement \cite{beta_VAE_ICLR_2017} results in the \emph{posterior collapse} \cite{ELBO_surgery_2016,posterior_collapse_ICLR_2019,posterior_collapse_Neurips_2019} i.e., uninformative latent variables leading to poor reconstruction. The use of additional regularization loss in the VAE objective \cite{TC_VAE_Neurips_2019,Factor_VAE_ICML_2018} for better disentanglement often results in the mismatch between the 
aggregate posterior distribution and the prior  \cite{rosca2018distribution,avae_saha_2023}. The mismatch leads to the generation of poor-quality samples due to the presence of \emph{pockets/holes} in the encoded distribution. DLVMs other than the VAE, such as the AAE\cite{makhzani2016adversarial}, WAE\cite{tolstikhin2017wasserstein}, GENs\cite{gens_saha_2022}, and the AVAE \cite{avae_saha_2023} do not suffer from the posterior collapse and the GENs\cite{gens_saha_2022}, and the methods in \cite{gens_saha_2022,avae_saha_2023} closely match the prior. However, matching the aggregate posterior to the prior in these models does not encourage the alignment of the latent generative factors identified by the models with the latent axes, unlike the VAE. Subsequently, resulting in poor performance under the existing disentanglement metrics \cite{beta_VAE_ICLR_2017,Factor_VAE_ICML_2018,TC_VAE_Neurips_2019,DCI_ICLR_2018,DIP_VAE_ICLR_2018}.

In this paper, we propose a method to \emph{evaluate} disentanglement of any trained DLVM, and not particularly VAEs. The proposed technique identifies directions (unit vectors) in the latent space representing latent variables associated with the true generative factors instead of relying on the cardinal latent axes as in VAEs. Existing metrics for disentanglement analysis use synthetic datasets with \emph{known latent factors} to evaluate the performance of any DLVM \cite{Dis_benchmark_ICML_2019} as the true latent factors are unknown for real-world datasets. We use the same labels to determine the latent directions (representing the latent factors) for any DLVM. The proposed technique for identifying latent vectors presents a generalized framework that results in better metric scores across DLVMs, particularly for methods that match aggregate posterior distributions, e.g., the AAE \cite{makhzani2016adversarial}, WAE \cite{tolstikhin2017wasserstein}, and AVAE \cite{avae_saha_2023}.

\section{Proposed method}
\subsection{Background}
The VAE use a probabilistic encoder, $\mathbf{E_{\phi}}$, and a probabilistic decoder, $\mathbf{D_{\theta}}$, to represent $q_{\phi}(\z{} \mid \x{})$ and, $p_{\theta}(\x{} \mid \z{})$, respectively. Both $\mathbf{E_{\phi}}$ and $\mathbf{D_{\theta}}$ are usually deep neural networks parameterized by $\phi$ and $\theta$, respectively. The prior distribution,  $p(\z{})=\Gauss{\zero}{\Id}$ and
the surrogate posterior is a factorized Gaussian distribution with diagonal covariance (assuming independent latent dimensions), which is defined as follows:
\begin{align}\label{eq:VAE_Surr_posterior}
q_{\phi}(\z{} \mid \x{}) = \Gauss{\boldsymbol\mu_{\x{}}}{\boldsymbol\sigma_{\x{}}^2\Id}, \text{where } \boldsymbol\mu_{\x{}}, \boldsymbol\sigma_{\x{}}^2 \leftarrow \mathbf{E_{\phi}}(\x{}).
\end{align}
The choice of the Gaussian distribution as the posterior, $q_{\phi}(\z{} \mid \x{})$, helps in efficient computation (reparameterization trick) of the $\mathbb{E}_{q_{\phi} (\z{} \mid \x{})} p_{\theta}(\x{} \mid \z{})$ in the VAE objective function (ELBO),
\begin{align}\label{eq:VAE_ELBO}
\max_{\theta, \phi} \mathbb{E}_{p(\x{})} \Bigl[ \mathbb{E}_{q_{\phi}(\z{} \mid \x{})} \log p_{\theta}(\x{} \mid \z{}) - \operatorname{KL} \Bigl( q_{\phi}(\z{} \mid \x{}) \lvert\rvert p(\z{}) \Bigr) \Bigr].
\end{align}
In contrast, DLVMs matching the aggregate posterior to the prior minimizes the $\operatorname{KL} \left( q_{\phi}(\z{}) \lvert\rvert p(\z{}) \right)$ in equation \ref{eq:VAE_ELBO} and uses a deterministic encoder-decoder.

\subsection{Discovery of the Latent Directions}\label{sec:Latent_directions}
\begin{algorithm}[htb]
\caption{: \textbf{Determine latent directions for generative factors in DLVMs, in a general setup}}
\hspace*{\algorithmicindent} \textbf{Input:} Trained encoder $E_\phi$, Latent factors $\mathcal{F}$, Values for the latent factors $\mathcal{V}$ ($\mathcal{V}_{i}$ has the values for the factor, $\mathcal{F}_{i}$), $L$ samples used for the PCA analysis of a generative factor ($\mathcal{F}_{i}$ is set to a fixed value chosen from $\mathcal{V}_{i}$ and all others factors, $\mathcal{F}_{-i}$ ,i.e., $\mathcal{F} \setminus i$, are allowed to vary), Number of PCA analysis ($N$) to determine the direction of the generative factor, $\mathcal{F}_{i}$.\\
\hspace*{\algorithmicindent} \textbf{Output:} Directions in the latent space, $\mathcal{D}$, for all the generative factors, $\mathcal{F}$.
\begin{algorithmic}[1]
\STATE $\mathcal{D} \gets \emptyset$
\FOR {Each ground truth factor $k$ in $\mathcal{F}$}
\STATE Sample $N$ values for the ground truth factor $k$, $\mathcal{S}^{k}$
\STATE $\mathcal{U} \gets \emptyset$
\FOR {each element $\mathcal{S}^{k}_{i}$ in $\mathcal{S}^{k}$}
\STATE Sample factors other than $k$ ($\mathcal{F}_{-k}$) $L$ times, $\mathcal{S}^{-k}$ \\ 
\COMMENT{Concatenate $\mathcal{S}^{k}_{i}$ to $L$ samples in $\mathcal{S}^{-k}$}
\STATE $\mathcal{S}^{i} \gets {\mathcal{S}^{-k}}^\frown \mathcal{S}^{k}_{i}$
\STATE Get observed data, $\mathcal{X}^{i}$, corresponding to $\mathcal{S}^{i}$, where $\mathcal{X}^{i}= \{x^{i}_{1}, x^{i}_{2} \ldots x^{i}_{L}\} \text{ and } x^{i}_{j} \in \mathbb{R}^{d}$
\STATE $\mathcal{Z}^{i} \gets E_\phi(\mathcal{X}^{i})$, where $\mathcal{Z}^{i} = \{z^{i}_{1}, z^{i}_{2} \ldots z^{i}_{L}\}$ and $z^{i}_{j} \in \mathbb{R}^{l}$
\STATE $\{(\sigma_{j}, u_{j})\}_{j=1}^{l} \gets \mathbb{PCA}(\mathcal{Z}^{i})$, where  $(\sigma_{j}, u_{j})$ represents the eigenvector ($u_{j}$) and the corresponding variance ($\sigma_{j}$) estimated from the PCA
\STATE $u_{i}$ is the eigenvector with the minimum variance
\STATE $\mathcal{U} \gets \mathcal{U} \cup u_{i}$
\ENDFOR\\
\COMMENT{\textbf{Comment: Estimate $u^{*}$ representing the factor, $\mathcal{F}_{i}$}} \\
\STATE $\hat{U} \gets 0$
\FOR {$u_{i}$ in $\mathcal{U}$}
\STATE $\hat{U} \gets \hat{U} + u_{i}u_{i}^{T}$
\ENDFOR
\STATE $\hat{U} \gets \frac{\hat{U}}{N}$
\STATE $\{(\sigma_{j}, u_{j})\}_{j=1}^{l} \gets \mathbb{EIGEN}(\hat{U})$
\STATE $u^{*}$ is the eigenvector with the maximum variance
\STATE $\mathcal{D} \gets \mathcal{D} \cup u^{*}$
\ENDFOR
\end{algorithmic}
\label{alg:Dis_PCA}
\end{algorithm}

In this method, we use the latent representations produced by the trained DLVMs to discover latent directions representing ground truth factors. Given the \emph{known} factors of variations, $\mathcal{F}$, for a dataset we use samples from the observed data corresponding to the $i$-th factor of variation, $\mathcal{F}_{i}$, to determine the direction in the latent space representing the latent variable for the factor, $\mathcal{F}_{i}$. To determine the direction corresponding to a factor $\mathcal{F}_{i}$, $L$ observed data are selected, where the factor $\mathcal{F}_{i}$ is fixed to an \emph{unique value} (chosen at random from $\mathcal{V}_{i}$ containing the possible values for the $i$-th latent factor) for all $L$ samples. The remaining factors, $\mathcal{F}_{-i}$, are assigned different values in each instance of the $L$ examples, resulting in $\mathcal{S}^{i}$ that is used to get the observed dataset, $\mathcal{X}^{i}$. The principal component analysis (PCA) is done on the encoding of $L$ observed data in $\mathcal{X}^{i}$, produced by a trained DLVM, and the eigenvector with minimum variance, $u_{i}$, is chosen as the representative of the ground truth factor, $\mathcal{F}_{i}$ (assigned a fixed value). This step is similar to the data generation technique of the FactorVAE metric \cite{Factor_VAE_ICML_2018} that associates a latent axis with a generative factor based on the variance along the latent axes. 

Determination of the minimum variance eigenvector is repeated multiple times ($N$) to capture the variation in $u_{i}$'s for different configurations of $\mathcal{F}_{i}$ and $\mathcal{F}_{-i}$. The optimum unit vector, $u^*$, representing the factor, $\mathcal{F}_{i}$, is obtained by solving the optimization problem $\max \sum_{i=1}^{N} (u^{*T}u_{i})^2$, where the $u_{i}$ is an eigenvector. The solution to the optimization problem is the eigendecomposition of the mean outer product of the eigenvectors ($u_{i}$), and $u^*$ is the eigenvector with the maximum variance. The above steps are repeated to get the latent directions ($u^*$) for all the ground truth factors. The outline of the algorithm is presented in Algorithm \ref{alg:Dis_PCA}.

\subsection{Evaluation}\label{sec:Evaluation}
In this work, we follow the strategy proposed in the FactorVAE metric \cite{Factor_VAE_ICML_2018} and MIG metric \cite{TC_VAE_Neurips_2019} to devise techniques for disentanglement analysis using the latent directions, $\mathcal{D}$. However, using the estimated latent directions is not limited to these metrics. We name the proposed metrics as the \emph{PCA FactorVAE metric} and the \emph{PCA MIG metric}. Finding latent directions (representing latent variables), $\mathcal{D}$, corresponding to generative factors, $\mathcal{F}$, is a generalization of the majority vote classifier used in the FactorVAE metric. Thus, we can use the latent directions $\mathcal{D}$ and $u_{i}$ (estimated using the inner loop in algorithm \ref{alg:Dis_PCA}) for different values of $\mathcal{F}_{i}$ to predict the latent factor, $\mathcal{\hat{F}}_{i}$. In the PCA FactorVAE metric, we use a similarity measure between the $u_{i}$ and the set of latent directions, $\mathcal{D}$, to predict the corresponding latent factor, $\mathcal{\hat{F}}_{i}$, and compare it to the true generative factor, $\mathcal{F}_{i}$, using cosine similarity measure (normalized correlation). The \emph{prediction accuracy} of a model is the score for the PCA FactorVAE metric. In the PCA MIG metric, latent representations $(\mathcal{Z}=E_\phi(\mathcal{X}))$ are projected onto the latent directions, $\mathcal{D}$, and the MIG of the transformed representations $(\mathcal{Z}'=\mathcal{Z}\mathcal{D}^T)$ gives the score.

\begin{table*}[htb]
    \centering
    \caption{Disentanglement scores of competing methods trained with $10$ different seeds for multiple datasets (higher is better). The \textbf{best} score is in \textbf{bold}, and the \underline{second best} score is \underline{{underlined}}. We indicate the improvement in the metric scores using the blue color and the drop with the red color. We observe the maximum improvement in the metric scores of the AVAE for both datasets.} \label{tab:Disentanglement_study}
    \resizebox{\textwidth}{!}
    {%
    \begin{tabular}{c c c c c c c c c c c c c}
    \hline
    Dataset
    &&Method
    &&FactorVAE\;\;$\uparrow$&PCA FactorVAE\;\;$\uparrow$ &Diff\;\;$\uparrow$
    &&MIG\;\;$\uparrow$&PCA MIG\;\;$\uparrow$ &Diff\;\;$\uparrow$
    &&MSE\;\;$\downarrow$\\ \hline
    \multirow{7}{*}{\textsc{DSprites}}
    &&VAE 
    && $\underline{64.78 \pm 8.05}$ &$75.56 \pm 7.21$ &\color{blue}$10.78$
    && $0.06 \pm 0.02$ &$0.14 \pm 0.04$ &\color{blue}$\underline{0.12}$
    && $3.68 \pm 0.58$ \\
    &&$\beta$-TCVAE 
    && $\mathbf{75.55 \pm 3.52}$ &$69.12 \pm 15.08$ &\color{red}$-6.43$
    && $\mathbf{0.20 \pm 0.06}$ &$\underline{0.18 \pm 0.13}$ &\color{red}$-0.02$
    && $6.39 \pm 2.05$ \\
    &&DIP-VAE-I
    &&$61.77 \pm 8.96$ &$70.68 \pm 6.89$ &\color{blue}$8.91$
    &&$\underline{0.13 \pm 0.07}$ &$0.12 \pm 0.06$ &\color{red}$-0.01$
    &&$3.61 \pm 0.47$ \\
    &&DIP-VAE-II 
    &&$60.70 \pm 10.97$ &$69.10 \pm 2.15$ &\color{blue}$8.40$
    &&$0.08 \pm 0.04$ &$0.09 \pm 0.02$ &\color{blue}$0.01$
    &&$3.46 \pm 0.33$ \\
    &&RAE 
    &&$64.21 \pm 6.69$ &$\mathbf{82.03 \pm 2.58}$ &\color{blue}$\underline{17.82}$
    &&$0.04 \pm 0.01$ &$0.16 \pm 0.03$ &\color{blue}$\underline{0.12}$
    && $\mathbf{2.51 \pm 0.20}$ \\
    &&WAE 
    && $47.87 \pm 5.90$ &$62.72 \pm 6.85$ &\color{blue}$14.85$ 
    && $0.02 \pm 0.01$ &$0.07 \pm 0.02$ &\color{blue}$0.05$
    && $3.69 \pm 0.34$ \\
    &&AVAE 
    && $59.03 \pm 2.62$ &$\underline{79.17 \pm 1.64}$ &\color{blue}$\mathbf{20.14}$ 
    && $0.02 \pm 0.00$ &$\mathbf{0.20 \pm 0.02}$ &\color{blue}$\mathbf{0.18}$
    && $\underline{2.98 \pm 0.28}$ \\
    [1.2ex] \hline
    \multirow{7}{*}{\textsc{3D Shapes}}
    &&VAE 
    && $56.19 \pm 8.85$ &$55.00 \pm 13.56$ &\color{red}$-1.19$
    && $0.13 \pm 0.13$ &$0.15 \pm 0.15$ &\color{blue}$0.02$
    && $10.47 \pm 1.10$ \\
    &&$\beta$-TCVAE 
    && $\mathbf{75.51 \pm 12.84}$ &$\underline{76.65 \pm 15.78}$ &\color{blue}$1.14$
    && $\mathbf{0.40 \pm 0.22}$ &$\underline{0.46 \pm 0.20}$ &\color{blue}$0.06$
    && $11.54 \pm 1.86$ \\
    &&DIP-VAE-I
    &&$51.94 \pm 1.91$ &$48.94 \pm 2.02$  &\color{red}$-2.00$
    &&$0.06 \pm 0.02$ &$0.23 \pm 0.03$ &\color{blue}$0.17$
    &&$\underline{10.16 \pm 0.83}$ \\
    &&DIP-VAE-II 
    &&$63.66 \pm 11.26$ &$66.04 \pm 18.15$ &\color{blue}$2.38$ 
    &&$\underline{0.24 \pm 0.18}$ &$0.27 \pm 0.19$ &\color{blue}$0.03$
    &&$12.10 \pm 2.64$ \\
    &&RAE 
    &&$53.57 \pm 13.14$ &${70.85 \pm 20.02}$ &\color{blue}$\underline{17.28}$
    &&${0.03 \pm 0.02}$ &${0.33 \pm 0.17}$ &\color{blue}$\underline{0.30}$
    && ${10.77 \pm 1.25}$ \\
    &&WAE 
    && $48.54 \pm 3.04$ &$52.34 \pm 3.0$ &\color{blue}$3.80$
    && $0.05 \pm 0.03$ &$0.20 \pm 0.05$ &\color{blue}$0.15$
    && $\mathbf{9.80 \pm 1.95}$ \\
    &&AVAE 
    && $\underline{72.90 \pm 7.30}$ &$\mathbf{91.93 \pm 3.27}$ &\color{blue}$\mathbf{19.03}$ 
    && $0.08 \pm 0.02$ &$\mathbf{0.67 \pm 0.04}$ &\color{blue}$\mathbf{0.59}$
    && ${10.29 \pm 0.37}$ \\
    [1.2ex] \hline
    \end{tabular}}
\end{table*}

\begin{figure*}
    \centering
    \includegraphics[width=1.0\textwidth]{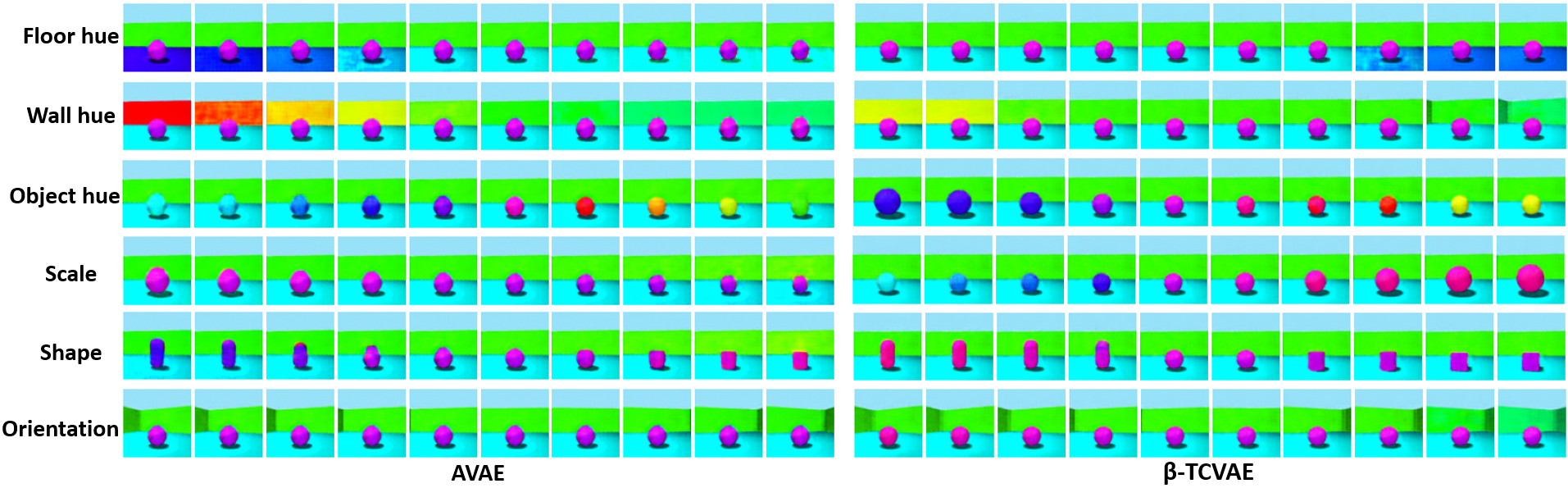}
    \caption{Latent traversal of the 3D Shapes dataset \cite{3dshapes18} in the range  $[-\sigma, \sigma]$ for models trained using the AVAE and $\beta$-TCVAE. The latent factors are mentioned in the left column. All latent factors are represented by independent latent variables in the AVAE, with almost no overlap between latent variables except a slight variation in object color with shapes. For the $\beta$-TCVAE, we observe the entanglement of the multiple latent factors, such as the object color with the scale and the wall color with orientation. The visualization justifies the \emph{improved} metrics scores of the AVAE in Table \ref{tab:Disentanglement_study} using the proposed evaluation method.}
    \label{fig:3DShapes_disentanglement}
\end{figure*}

\begin{figure*}
    \centering
    \includegraphics[width=1.0\textwidth]{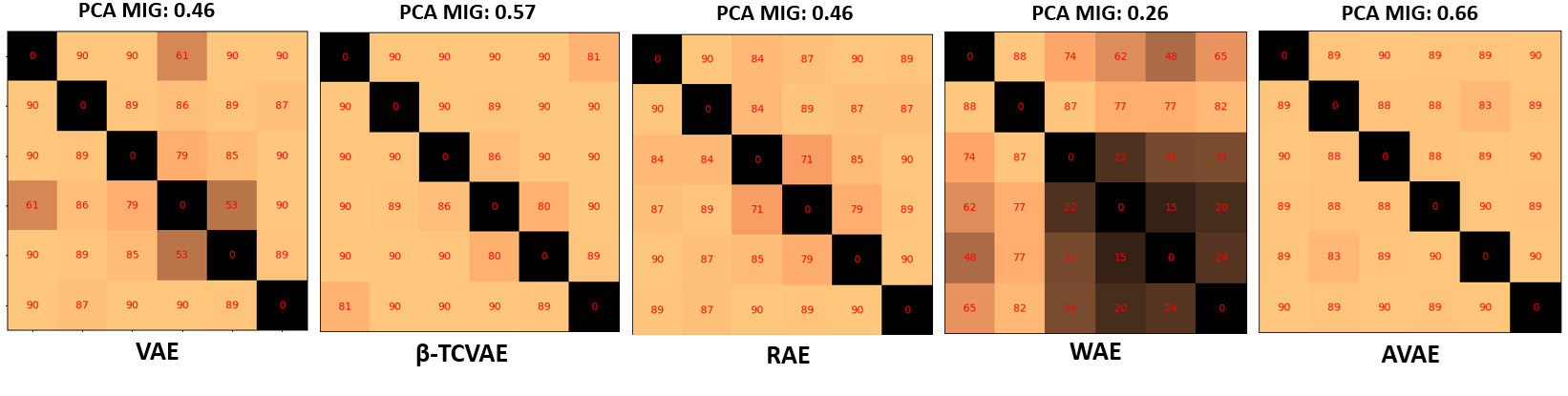}
    \caption{Pairwise angle between the latent directions (six directions) estimated by Algorithm \ref{alg:Dis_PCA} for different DLVMs using the corresponding latent representations produced for the 3D Shapes dataset \cite{3dshapes18}. The latent directions should be orthogonal to each other for better disentanglement. Deviation from the orthogonality indicates entanglement of the ground truth generative factors that result in poor metric scores, as observed in the WAE. To interpret the estimated latent directions in the analysis of disentanglement, we report the corresponding PCA MIG metric score for each model.}
    \label{fig:latent_directions}
\end{figure*}

\section{Experiments}
\subsection{Benchmark Methods \& Datasets}
Other than the regular VAE \cite{kingma2014auto}, we consider different variations of the VAE that modify the original formulation to match the aggregate posterior to the prior, such as the FactorVAE \cite{Factor_VAE_ICML_2018}, and $\beta$-TCVAE \cite{TC_VAE_Neurips_2019}, for comparison. We do not consider the FactorVAE as it uses a discriminator to optimize the objective function, which is challenging to train. For the same reason, we do not study the AAE \cite{makhzani2016adversarial} in this work. We study the DIP-VAE \cite{DIP_VAE_ICLR_2018} that adds a regularizer to the VAE objective function to better match the aggregate posterior to the prior. We also evaluate the RAE \cite{RAE} and AVAE \cite{avae_saha_2023} in this work that uses a deterministic encoder, unlike the regular VAE. The AVAE is a new method based on the formulation of the VAE that addresses the posterior collapse and closely matches the aggregate posterior. Other than the variants of the VAE, we consider the WAE  (with IMQ kernel) \cite{tolstikhin2017wasserstein} that matches aggregate posterior in the latent space.

We use the DSprites \cite{dsprites17} and 3D Shapes \cite{3dshapes18} datasets to evaluate DLVMs. The true generative factors of the observed data are known for both datasets. Annotated data is \emph{required} for the quantification of the disentanglement in the latent representations of trained DLVMs.

\begin{table}
\centering
\caption{Optimization settings for different methods.} \label{tab:hyperparameters}
\resizebox{0.49\textwidth}{!}
{%
\begin{tabular}{c c c c c c c}
\hline
Method&&Parameters&&{DSprites}&&{3D Shapes}\\ \hline
\textsc{$\beta$-TCVAE} && $\beta$: && $5$ && $5$\\
\textsc{DIP-VAE-I} && $(\lambda_{od}, \lambda_{d})$: && $(10, 100)$ && $(10, 100)$ \\
\textsc{DIP-VAE-II} && $(\lambda_{od}, \lambda_{d})$: && $(10, 10)$ && $(10, 10)$ \\
\textsc{RAE} && $\beta$: && $1e-04$ && $1e-04$ \\
\textsc{RAE} && \textsc{Dec-L}$2$-\textsc{reg}: && $1e-06$ && $1e-06$\\
\textsc{WAE} && \textsc{Recons-scalar}: && $0.05$ && $0.05$\\
\textsc{WAE} && $\beta$: && $10$ && $10$ \\
\textsc{AVAE} && KDE samples $(m)$: && $10K$ && $10K$\\
\hline
\end{tabular}
}
\end{table}

\subsection{Implementation Details} For a given dataset, we use the same latent dimension, encoder-decoder architecture (as used in \cite{Dis_benchmark_ICML_2019}), and optimization strategies (such as the learning rate, learning rate scheduler, epochs, and batch size) for all the competing methods to ensure a fair comparison. We leverage the information of the known latent factors of the DSprites and 3D Shapes datasets to set the latent size as $l=6$ for both datasets. For the AVAE, the number of the KDE samples used is $10K$ for both the DSprites and 3D Shapes datasets. We run all methods with $10$ different seeds (producing different initialization) for the DSprites and 3D Shapes datasets. The objective function of several methods studied in this work has hyperparameters related to the regularization losses tuned for different datasets. Mostly, we have used hyperparameter settings suggested by the author or recommended in the literature\cite{TC_VAE_Neurips_2019, Dis_benchmark_ICML_2019}. The hyperparameters of the methods are reported in Table \ref{tab:hyperparameters}.

\subsection{Results}
The performance of the competing methods under the proposed disentanglement metrics is reported in Table \ref{tab:Disentanglement_study} for the DSprites \cite{dsprites17} and 3D Shapes \cite{3dshapes18} datasets. A relatively poor reconstruction loss indicates stronger regularization of the latent representation, possibly leading to better disentanglement. Therefore, knowing the reconstruction loss of DLVMs on different datasets is informative. In an ideal scenario, we expect a higher disentanglement score with low reconstruction loss. In this experiment, we consider the FactorVAE metric \cite{Factor_VAE_ICML_2018} and the MIG metric \cite{TC_VAE_Neurips_2019} as the baseline to demonstrate the effectiveness of the proposed evaluation technique in Section \ref{sec:Evaluation} using the latent directions, $\mathcal{D}$, estimated by Algorithm \ref{alg:Dis_PCA}.

Comparing the PCA MIG metric with the MIG metric \cite{TC_VAE_Neurips_2019} (baseline) demonstrates the impact of using the latent directions, $\mathcal{D}$, as latent variables relative to the latent axes. We observe an overall improvement in the performance of all the methods studied in this work using the latent directions, $\mathcal{D}$, in the computation of the MIG scores, but the marginal drop on the DSprites dataset for the $\beta$-TCVAE and DIP-VAE-I. Likewise, except for the drop in the performance of the $\beta$-TCVAE on the DSprites dataset using the estimated latent directions, we observe a consistent improvement in the performance of all the methods. Methods matching the aggregate distribution benefit the most using the latent directions, $\mathcal{D}$.

Considering the metric scores reported in Table \ref{tab:Disentanglement_study}, the AVAE produces the best score under both the metrics for the 3D Shapes dataset with a slightly higher MSE score (second best). This indicates that the AVAE achieves better disentanglement without compromising the quality of the reconstructed data, a desired property of a DLVM. The AVAE achieves significantly higher scores than the $\beta$-TCAVE for both the metrics and sets a new SOTA result for the 3D Shapes dataset. Higher metric scores for the AVAE are corroborated by the latent traversal along the latent directions, $\mathcal{D}$, shown in Figure \ref{fig:3DShapes_disentanglement}. Overall, the performance of the AVAE is consistent under both metrics relative to the competing methods for both datasets.

The latent directions estimated by Algorithm \ref{alg:Dis_PCA} should be orthogonal for independent latent variables. The collapse of the latent directions or small angle between them indicates the entanglement of ground truth generative factors to latent variables that should affect the disentanglement scores. Figure \ref{fig:latent_directions} shows the angles between the estimated latent directions for different DLVMs trained on the 3D Shapes dataset. Every latent direction estimated for the AVAE is almost perpendicular to all other latent directions. This property explains the high metric scores of the AVAE in Table \ref{tab:Disentanglement_study} on both datasets.

\section{Conclusion}
We present a statistical method to \emph{evaluate disentanglement} in trained DLVMs using the estimated latent directions representing the generative factors of a dataset that \emph{can be different from the cardinal latent axes}. We demonstrate improvements in metric scores for methods matching aggregate posterior distributions, such as the AVAE \cite{avae_saha_2023}. Therefore, we show limitations in the existing metrics that rely on cardinal latent axes representing the generative factors. This work is supported by the National Institutes of Health grant R01ES032810.

\bibliographystyle{IEEEtran}
\bibliography{paper}

\end{document}